\documentclass{sysu_preprint}
\DeclareTextFontCommand{\textbf}{\bfseries}
\usepackage{fix-cm}
\usepackage{amsmath,amssymb,amsthm,mathtools}
\usepackage{graphicx,booktabs,multirow,tabularx,array,colortbl,makecell}
\usepackage{algorithm,algorithmic,pifont,float,pdfpages}

\newcolumntype{C}{>{\centering\arraybackslash}p{1.7cm}}
\definecolor{lightblue}{RGB}{200,220,255}
\definecolor{lightyellow}{RGB}{255,255,200}
\definecolor{lightgreen}{RGB}{220,255,220}
\definecolor{modulecolor}{RGB}{0,100,255}
\definecolor{highlight}{RGB}{255,100,0}

\hypersetup{pdftitle={OGPO: One-Step Generative Policy Optimization for Real-Time Robot Control},pdfauthor={Guowei Zou, Haitao Wang, Hejun Wu, Yukun Qian, Yuhang Wang, Weibing Li}}
\title{OGPO: One-Step Generative Policy Optimization for Real-Time Robot Control}
\renewcommand{\authorlist}{\authorfont{\sffamily Guowei Zou\textsuperscript{*}, Haitao Wang\textsuperscript{*}, Hejun Wu\textsuperscript{\textdagger}, Yukun Qian, Yuhang Wang, Weibing Li\textsuperscript{\textdagger}}}
\abstract{Real-time robot control demands fast action generation. Diffusion and flow matching policies for robot control require multi-step sampling, limiting their deployment in real-time scenarios. Natively reducing the sampling steps to one sacrifices representation quality and task performance, creating a trilemma among speed, fidelity, and performance. We present One-Step Generative Policy Optimization (OGPO), a systematic framework to resolve this trilemma. OGPO first pairs a lightweight architecture with the interval velocity principle for distillation-free one-step inference, while representation spreading prevents representation quality degradation. It then performs on-policy reinforcement learning (RL) fine-tuning on this fast, stable policy to break the imitation learning ceiling. Experiments on RoboMimic and OpenAI Gym benchmarks show that OGPO matches or exceeds multi-step baselines while achieving 5--20$\times$ inference speedup and over 120Hz control frequency. Physical deployment on a Franka-Emika-Panda robot validates real-world applicability.}
\newcommand{\teaserfigure}{\begin{center}\centering
\includegraphics[width=.98\linewidth]{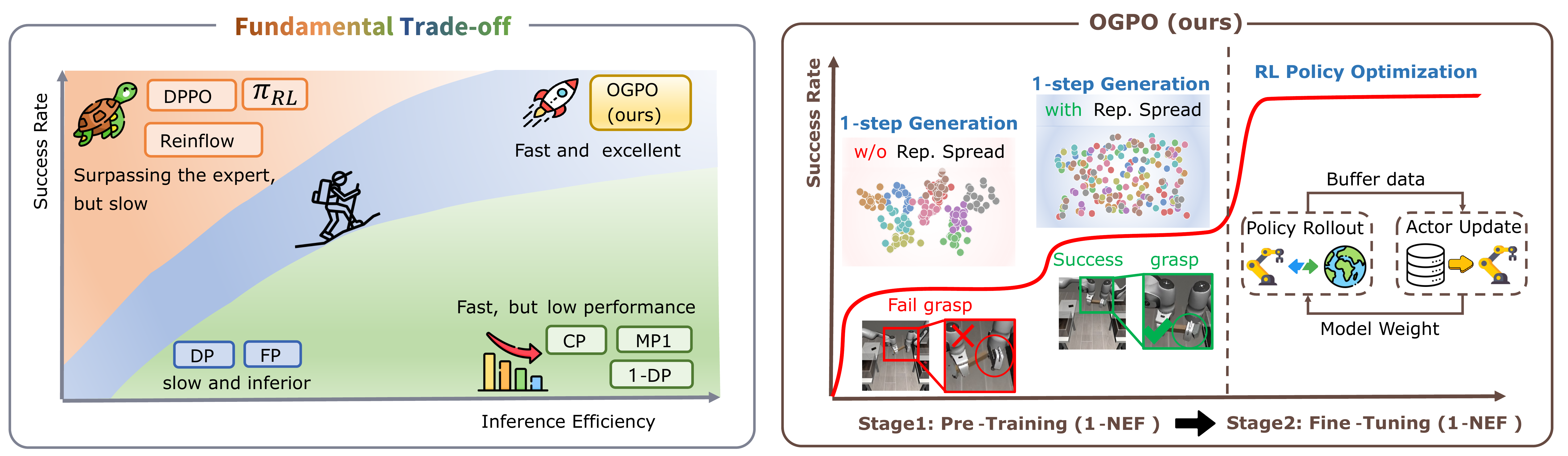}
\captionsetup{hypcap=false}\captionof{figure}{\textbf{Left:} Multi-step methods (DPPO, ReinFlow) are accurate but slow; one-step methods (CP, MP1) are fast but underperform. OGPO occupies the upper-right region. \textbf{Right:} Stage~1 trains a CVF with representation spreading for one-step inference; Stage~2 applies RL fine-tuning to surpass expert demonstrations.}
\label{fig:motivation}\end{center}}
\definecolor{coverpurple}{HTML}{49308C}
\definecolor{covergray}{HTML}{F1F4F8}
\renewcommand{\maketitle}{%
  \thispagestyle{plain}%
  \begin{tcolorbox}[enhanced,colback=covergray,frame hidden,arc=3mm,
    left=6mm,right=6mm,top=4mm,bottom=4mm,boxsep=0pt,before skip=0pt,after skip=5mm]
    {\raggedright\sffamily\bfseries\color{coverpurple}\fontsize{17}{20}\selectfont
      \titlelist\par}
    \vspace{2mm}
    {\raggedright\renewcommand{\authorfont}{\fontsize{10}{12}\selectfont}
      \bfseries\authorlist\par}
    \vspace{2mm}
    {\raggedright\sffamily\bfseries\color{coverpurple}\fontsize{11}{13}\selectfont Sun Yat-sen University\par}
    {\sffamily\fontsize{8}{10}\selectfont $^{*}$Equal contribution. $^{\dagger}$Corresponding authors.\par}\vspace{3mm}
    {\color{coverpurple!30}\hrule height 0.3pt}
    \vspace{1mm}
    \teaserfigure
    \vspace{1mm}
    {\color{coverpurple!30}\hrule height 0.3pt}
    \vspace{2mm}
    {\small\hypersetup{urlcolor=coverpurple}
      {\sffamily\bfseries Project Page:} \url{https://ogpo-project.github.io/}\par
      {\sffamily\bfseries Code:} \url{https://github.com/ogpo-project/OGPO}\par
      {\sffamily\bfseries Models:} \url{https://huggingface.co/ogpo-project/OGPO-checkpoints}\par
      {\sffamily\bfseries Datasets:} \url{https://huggingface.co/datasets/ogpo-project/OGPO-datasets}\par
    }

  \end{tcolorbox}
  {\centering\sffamily\bfseries Abstract\par}
  \vspace{2mm}
  {\setlength{\parindent}{0pt}\abstractlist\par}
  \par\vspace{3mm}
}

\begin{document}
\raggedbottom
\maketitle
\section{Introduction}

Real-time control is essential for dexterous robotic manipulation~\cite{shao2025vlasurvey,li2025worldmodelsurvey,yang2026worldmodels,mei2026videogenrobotics}. Generative models based on diffusion and flow matching have achieved remarkable success in modeling complex action distributions~\cite{diffusionpolicy2023,dp3_2023,flowpolicy2024}. Yet their multi-step sampling incurs significant latency, fundamentally limiting deployment in time-critical scenarios. Existing one-step methods such as Consistency Policy~\cite{consistencypolicy2024} and 1-DP~\cite{onedp2025} rely on distillation from multi-step teachers, inheriting complex pipelines and bounded by teacher's performance. Meanwhile, RL fine-tuning methods~\cite{dppo2024,reinflow2024,pirl2025,wovr2026,worldgymnast2026} can surpass expert demonstrations but all operate on slow multi-step models. Naively reducing to one step sacrifices representation quality and task performance, creating a trilemma among \emph{speed}, \emph{fidelity}, and \emph{performance}, as illustrated in Fig.~\ref{fig:motivation}. In this work, we aim to improve one-step generative methods to enhance real-time speed and performance for robot control.

Two coupled bottlenecks underlie this trilemma. The first is \emph{representation quality degradation} in one-step generation. Current generative policies place no explicit constraint on encoder representations~\cite{repa2024,soda2023}. Multi-step models implicitly compensate through iterative error correction across denoising steps~\cite{implicit_reg_diffusion2025}. When generation is reduced to a single step, this correction mechanism vanishes entirely. Any information lost in one forward pass is irrecoverable, directly degrading action quality. The second is the \emph{imitation learning ceiling}~\cite{simchowitz2025compounding,beyondexpert2025,rile2025,resfit2025}. All existing one-step policies are bounded by their supervision source. Distillation-based methods~\cite{consistencypolicy2024,onedp2025} cannot exceed the teacher. Native one-step models~\cite{mp1_meanflow} cannot exceed the expert dataset. Online RL offers a principled way to break this ceiling. These two bottlenecks reinforce each other, making the joint achievement of speed, fidelity, and performance elusive. No prior work has resolved this trilemma.

We propose One-Step Generative Policy Optimization (\textbf{OGPO}), a systematic framework to resolve this trilemma. OGPO first designs a \emph{Compact Velocity Field} (CVF) that pairs a lightweight architecture with the interval velocity principle for distillation-free one-step inference. To counter representation quality degradation, it applies \emph{representation spreading}, a plug-and-play regularization that preserves discriminative encoder features. OGPO then performs on-policy RL fine-tuning on this fast, stable one-step policy. One-step inference accelerates rollout collection. Stable representations support reliable value estimation. Together, they enable RL to surpass expert demonstrations while maintaining real-time speed (Fig.~\ref{fig:motivation}).

Our contributions are:
(1) \textbf{Finding:} We identify representation quality degradation as the critical bottleneck preventing online RL optimization of one-step generative policies. We show that contrastive regularization on encoder features, which is merely a performance enhancement for multi-step models, becomes structurally necessary in the one-step setting.
(2) \textbf{Framework:} Based on this finding, we present OGPO, a two-stage pipeline that first trains a stable one-step policy and then optimizes it online. OGPO achieves over 120Hz control with 5--20$\times$ speedup while matching or exceeding multi-step baselines.
(3) \textbf{Validation:} We validate OGPO on RoboMimic~\cite{robomimic2021}, OpenAI Gym~\cite{gym2016}, and a physical Franka-Emika-Panda robot, establishing state-of-the-art across both manipulation and locomotion tasks.

\section{Related Work}


\subsection{Multi-Step Generative Policies}

Diffusion Policy~\cite{diffusionpolicy2023} pioneered visuomotor control via denoising diffusion. DP3~\cite{dp3_2023} extended it to 3D inputs. Flow matching~\cite{lipman2023flow,boffi2025flowmap} and Rectified Flow~\cite{rectifiedflow2022} offer alternatives through learned velocity fields, applied in Flow Policy~\cite{flowpolicy2024}, shortcut flows~\cite{shortcutflow2025}, and foundation models $\pi_0$~\cite{pi0_2024}, $\pi_{0.5}$~\cite{pi05_2024}. Vision-Language-Action models such as OpenVLA~\cite{openvla2024}, Octo~\cite{octo2024}, RDT-1B~\cite{rdt2024}, and GR-2~\cite{gr2_2024} further scale generative policies to diverse embodiments and tasks. The challenge of efficient diffusion-based generation has also been explored in other multimedia domains, including motion synthesis~\cite{stablemofusion2024} and audio generation~\cite{audiolcm2024}. Despite their expressiveness, these methods require tens to hundreds of network function evaluations (NFEs) per action and cannot surpass expert performance. OGPO addresses both limitations through one-step generation and RL fine-tuning.

\subsection{One-Step Generation Methods}

CP~\cite{consistencypolicy2024} and 1-DP~\cite{onedp2025} reduce generation to one step via distillation but remain bounded by teacher's capability~\cite{manicm2025,maniflow2025,twostep2025,chen2025piflow,liu2025scott}. Similar consistency-based acceleration has been applied in speech synthesis~\cite{comospeech2023}. MP1~\cite{mp1_meanflow} first applies averaged velocity fields~\cite{meanflow2025,improved_meanflow2025,splitmeanflow2025,meanflow_transformer2025} for distillation-free one-step generation. However, it does not address representation quality degradation or the imitation learning ceiling, and uses a heavy UNet backbone (255M params) with 3D point cloud input. Representation spreading was proposed by \citet{diffuse_and_disperse} for multi-step diffusion models. D$^2$PPO~\cite{d2ppo} applied it to multi-step diffusion policy fine-tuning. In both cases, it serves as a \emph{performance enhancement} for models that already function without it. Our key insight is that in the one-step setting, this regularization transitions from optional to \emph{structurally necessary}: without iterative error correction, representation quality becomes the sole bottleneck for action fidelity.

\subsection{RL Fine-tuning for Generative Policies}

Recent work has explored RL fine-tuning of generative policies to surpass expert demonstrations~\cite{dppo2024,reinflow2024,pirl2025}. $\pi_{RL}$~\cite{pirl2025} further scaled this paradigm to Vision-Language-Action models~\cite{lu2025vlarl,liu2025flowgrpo,mcallister2025fmpg}. WoVR~\cite{wovr2026} leverages world models as reliable simulators for post-training VLA policies with RL. World-Gymnast~\cite{worldgymnast2026} trains robots with RL inside a learned world model. However, all prior methods operate on multi-step policies with 20 or more NFEs, inheriting significant computational overhead during rollout collection~\cite{d2ppo,wagenmaker2025steering}. \textbf{OGPO is the first to demonstrate effective RL fine-tuning at $K{=}1$}. We derive a novel single-step policy gradient formulation that reduces variance and accelerates rollout by an order of magnitude. OGPO retains the one-step policy during both rollout collection and deployment, so its online optimization targets the same low-latency policy class used at test time. This alignment allows reward-driven refinement without reverting to a slow multi-step generator.

\FloatBarrier
\section{Methodology}
\suppressfloats[t]

\begin{figure}[!htbp]
\centering
\includegraphics[width=0.95\textwidth]{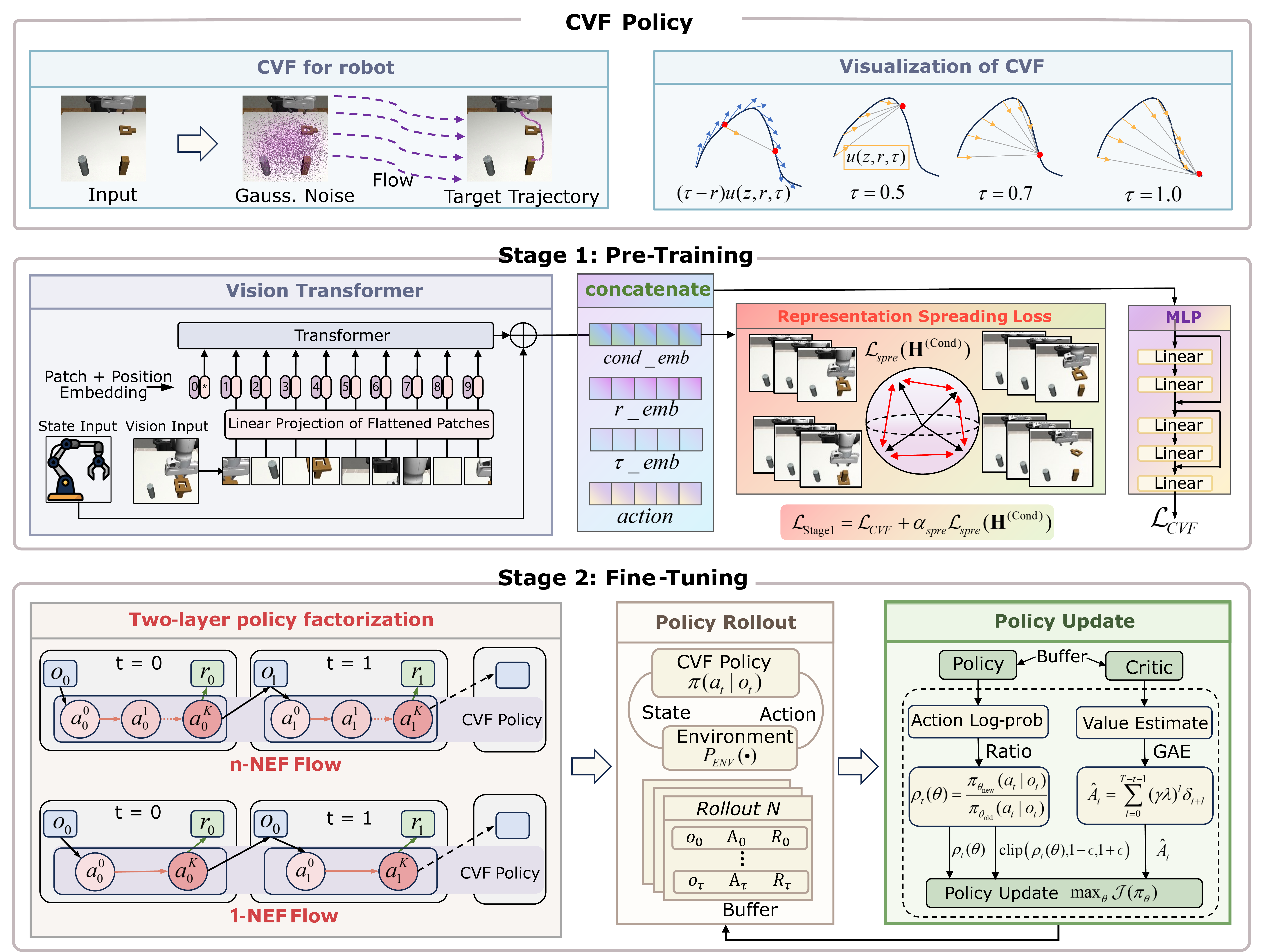}
\caption{\textbf{OGPO Framework Overview.} \textbf{Stage 1 (Top \& Middle):} Pre-training uses a Compact Velocity Field (CVF) for one-step inference, combined with representation spreading to prevent representation quality degradation. \textbf{Stage 2 (Bottom):} On-policy RL fine-tuning, formulated as a two-layer policy factorization: the outer layer is the environment Markov decision process (MDP), while the inner latent chain reparameterizes the action distribution.}
\label{fig:ogpo_framework}
\end{figure}

OGPO is a two-stage framework (Figure~\ref{fig:ogpo_framework}). Stage~1 trains a one-step policy that is both fast and representationally stable. Stage~2 optimizes this policy online to surpass expert demonstrations.

\subsection{Stage 1: One-Step Policy Training}
\label{sec:stage1}

\textbf{Architecture.} We use a single-layer Vision Transformer (ViT) with 128-dimensional embeddings and a multi-layer perceptron (MLP) action head, totaling 1.78M parameters. This compact design achieves over 120Hz inference.

\textbf{Velocity formulation.} The policy learns an averaged velocity field $u_\theta$ over arbitrary time intervals $[r, \tau]$~\citep{meanflow2025,improved_meanflow2025}. Given linear interpolation $z_\tau = (1-\tau)a + \tau\epsilon$ between target action $a$ (at $\tau{=}0$) and noise $\epsilon \sim \mathcal{N}(0,I)$ (at $\tau{=}1$), the averaged velocity satisfies:
\begin{equation}
    u(z_\tau,r,\tau) = v(z_\tau,\tau) - (\tau-r)\frac{d}{d\tau}u(z_\tau,r,\tau),
\end{equation}
where $v_\tau = \epsilon - a$ is the instantaneous velocity. At inference, a single forward pass produces the action: $a = z_1 - u_\theta(z_1, 0, 1, o)$. The velocity prediction loss is:
\begin{equation}
\mathcal{L}_{\text{CVF}} = \mathbb{E}_{(o,a),\,\epsilon,\,(r,\tau)} \left[
\big\| u_\theta(z_\tau, r, \tau, o) - u_{\text{tgt}} \big\|^2
\right],
\end{equation}
where $u_{\text{tgt}} = v_\tau - (\tau-r)\frac{d}{d\tau}u_\theta$.

\textbf{Representation spreading.} The velocity loss alone does not constrain the encoder representation structure. In multi-step models, iterative denoising implicitly regularizes representations. In the one-step setting, this implicit regularization vanishes, and representation quality degradation becomes the primary failure mode. To prevent this, we add a contrastive objective on the encoder conditional embedding $\mathbf{H}^{(\text{Cond})}$. We use an InfoNCE-L2 loss~\citep{diffuse_and_disperse} that maximizes pairwise distances:
\begin{equation}
\mathcal{L}_{\text{spre}} = -\mathbb{E} \left[ \log \frac{\exp(\|\mathbf{h}_i\|_2^2 / \tau)}{\sum_{k \neq i} \exp(-\|\mathbf{h}_i - \mathbf{h}_k\|_2^2 / \tau)} \right]
\end{equation}
The complete Stage~1 objective is:
\begin{equation}
\mathcal{L}_{\text{Stage1}} = \mathcal{L}_{\text{CVF}} + \alpha_{\text{spre}} \mathcal{L}_{\text{spre}}(\mathbf{H}^{(\text{Cond})}),
\end{equation}

where $\alpha_{\text{spre}} > 0$ balances velocity prediction and representation diversity. This joint training produces a one-step policy that is both fast and representationally rich, serving as the foundation for Stage~2.

\subsection{Stage 2: Online Policy Optimization}
\label{sec:stage2}

Stage~1 produces a strong one-step policy, but its performance is bounded by the expert data. Stage~2 breaks this ceiling through online reinforcement learning. The key enabler is that one-step inference is fast enough to collect on-policy rollouts efficiently, while the stable representations from Stage~1 support reliable value estimation.

\textbf{Policy formulation.} We formulate one-step generation as a $K$-step Markov chain ($K{=}1$ by default) with stochastic transitions:
\begin{equation}
a^{k+1} \sim \mathcal{N}\!\left(a^k - \Delta \tau_k \cdot u_\theta(a^k, \tau_{k+1}, \tau_k, o),\; \sigma^2 I\right)
\end{equation}
where $\tau_k = 1 - k/K$ and $\Delta \tau_k = 1/K$. The joint log-probability decomposes as:
\begin{equation}
\log \pi_\theta(a^{0:K} | o) = \log p_0(a^0) + \sum_{k=0}^{K-1} \log \mathcal{N}(a^{k+1} | \mu_k, \sigma^2 I)
\end{equation}
where $\mu_k = a^k - \Delta \tau_k \cdot u_\theta(a^k, \tau_{k+1}, \tau_k, o)$. For $K{=}1$, this reduces to a single Gaussian step from noise to action, yielding low-variance policy gradients.

\textbf{Optimization.} We optimize with Proximal Policy Optimization (PPO)~\cite{ppo2017} using clipped policy gradients, combined with behavior cloning (BC) regularization to prevent catastrophic forgetting:
\begin{equation}
\mathcal{L}_{\text{Stage2}} = \mathcal{L}_{\text{PPO}} + \lambda_{\text{BC}} \mathcal{L}_{\text{BC}}, \quad \mathcal{L}_{\text{BC}} = \mathbb{E}_{o}\left[\left\| a_{\omega}(o) - a_\theta(o) \right\|_2^2\right]
\end{equation}
where $a_\omega$ is the frozen pre-trained policy and $\lambda_{\text{BC}}$ decays during training. BC regularization is widely used in offline-to-online RL~\cite{td3bc2021,bc_reg2020}. We choose on-policy optimization over off-policy alternatives such as SAC and TD3 because the clipped objective naturally accommodates the log-probability factorization, on-policy rollouts directly leverage fast one-step inference, and trust-region constraints stabilize updates from a pre-trained initialization.

\begin{figure}[!htbp]
\centering
\begin{minipage}[t]{0.48\textwidth}
\begin{algorithm}[H]
\caption{OGPO Stage 1: One-Step Policy Training}
\label{alg:stage1}
\begin{algorithmic}[1]
\REQUIRE Dataset $\mathcal{D} = \{(o, a)\}$, weight $\alpha_{\text{spre}}$
\ENSURE Trained policy $\pi_\theta$
\FOR{each batch $\{(o^{(i)}, a^{(i)})\}_{i=1}^B$}
    \STATE Sample $\epsilon^{(i)} \sim \mathcal{N}(0, I)$, $(r^{(i)}, \tau^{(i)})$ from logit-normal
    \STATE $z_\tau^{(i)} = (1-\tau^{(i)})a^{(i)} + \tau^{(i)}\epsilon^{(i)}$
    \STATE $\hat{u}^{(i)} = u_\theta(z_\tau^{(i)}, r^{(i)}, \tau^{(i)}, o^{(i)})$
    \STATE $u_{\text{tgt}}^{(i)} = (\epsilon^{(i)} - a^{(i)}) - (\tau^{(i)}-r^{(i)})\frac{d}{d\tau}u_\theta$
    \STATE $\mathbf{h}^{(i)} = \phi_\theta(o^{(i)})$ \COMMENT{encoder output}
    \STATE $\mathcal{L}_{\text{CVF}} = \frac{1}{B}\sum_{i} \|\hat{u}^{(i)} - \text{sg}(u_{\text{tgt}}^{(i)})\|^2$
    \STATE $\mathcal{L}_{\text{spre}} = \text{InfoNCE-L2}(\{\mathbf{h}^{(i)}\})$
    \STATE $\theta \leftarrow \theta - \eta \nabla_\theta (\mathcal{L}_{\text{CVF}} + \alpha_{\text{spre}} \mathcal{L}_{\text{spre}})$
\ENDFOR
\end{algorithmic}
\end{algorithm}
\end{minipage}
\hfill
\begin{minipage}[t]{0.48\textwidth}
\begin{algorithm}[H]
\caption{OGPO Stage 2: Online Policy Optimization}
\label{alg:stage2}
\begin{algorithmic}[1]
\REQUIRE Pre-trained $\pi_\theta$, environment, $\lambda_{\text{BC}}$
\ENSURE Fine-tuned policy $\pi_\theta$
\FOR{iteration $n = 1, \ldots, N$}
    \STATE Collect rollouts with $K$-step sampling ($K{=}1$)
    \STATE Compute advantages $\hat{A}_t$ (GAE) and returns $R_t$
    \FOR{each PPO epoch}
        \STATE $\rho_t = \pi_\theta(a_t | o_t) / \pi_{\text{old}}(a_t | o_t)$
        \STATE $\mathcal{L}_{\text{PG}} = \mathbb{E}_t[\max(-\hat{A}_t \rho_t, -\hat{A}_t \text{clip}(\rho_t, 1{\pm}\epsilon))]$
        \STATE $\mathcal{L}_V = \frac{1}{2}\mathbb{E}_t[(V_\theta(o_t) - R_t)^2]$
        \STATE $\mathcal{L}_{\text{BC}} = \mathbb{E}_t[\|a_\omega(o_t) - a_\theta(o_t)\|^2]$
        \STATE $\theta \leftarrow \theta - \eta \nabla_\theta (\mathcal{L}_{\text{PG}} + \lambda_V \mathcal{L}_V + \lambda_{\text{BC}}(n) \mathcal{L}_{\text{BC}})$
    \ENDFOR
\ENDFOR
\end{algorithmic}
\end{algorithm}
\end{minipage}
\end{figure}

\FloatBarrier
\section{Experiments}
\suppressfloats[t]

We evaluate pre-training and fine-tuning on four RoboMimic manipulation tasks~\citep{robomimic2021} and OpenAI Gym locomotion tasks~\citep{gym2016}. All results are averaged over 3 random seeds with 100 evaluation episodes per seed. OGPO uses $K{=}1$ (one-step generation) for both training and inference unless otherwise noted; we additionally report $K{=}2$ and $K{=}5$ inference results to show multi-step compatibility. To ensure fair comparison, we adopt the same Simplified RoboMimic MH dataset used in prior work~\citep{dppo2024,reinflow2024}. For RoboMimic, we report task success rate; for Gym, we report episode reward; and for Kitchen, we report task completion rate. We fix the number of denoising steps for each evaluated curve, so differences in the plots reflect policy performance rather than a changing sampling budget.

Our experiments are designed to answer three research questions:

\noindent\textcolor{blue}{\textbf{RQ1:} Does OGPO address the representation quality degradation caused by the lack of iterative correction in one-step generation?}

\noindent\textcolor{blue}{\textbf{RQ2:} Does OGPO enable one-step generative policies to surpass the expert demonstration ceiling?}

\noindent\textcolor{blue}{\textbf{RQ3:} Does OGPO achieve better performance and faster inference than existing multi-step and one-step methods?}

\subsection{Representation Spreading Ablation (RQ1)}

\subsubsection{RoboMimic pre-training tasks}

\begin{figure}[!htbp]
\centering
\includegraphics[width=0.95\textwidth]{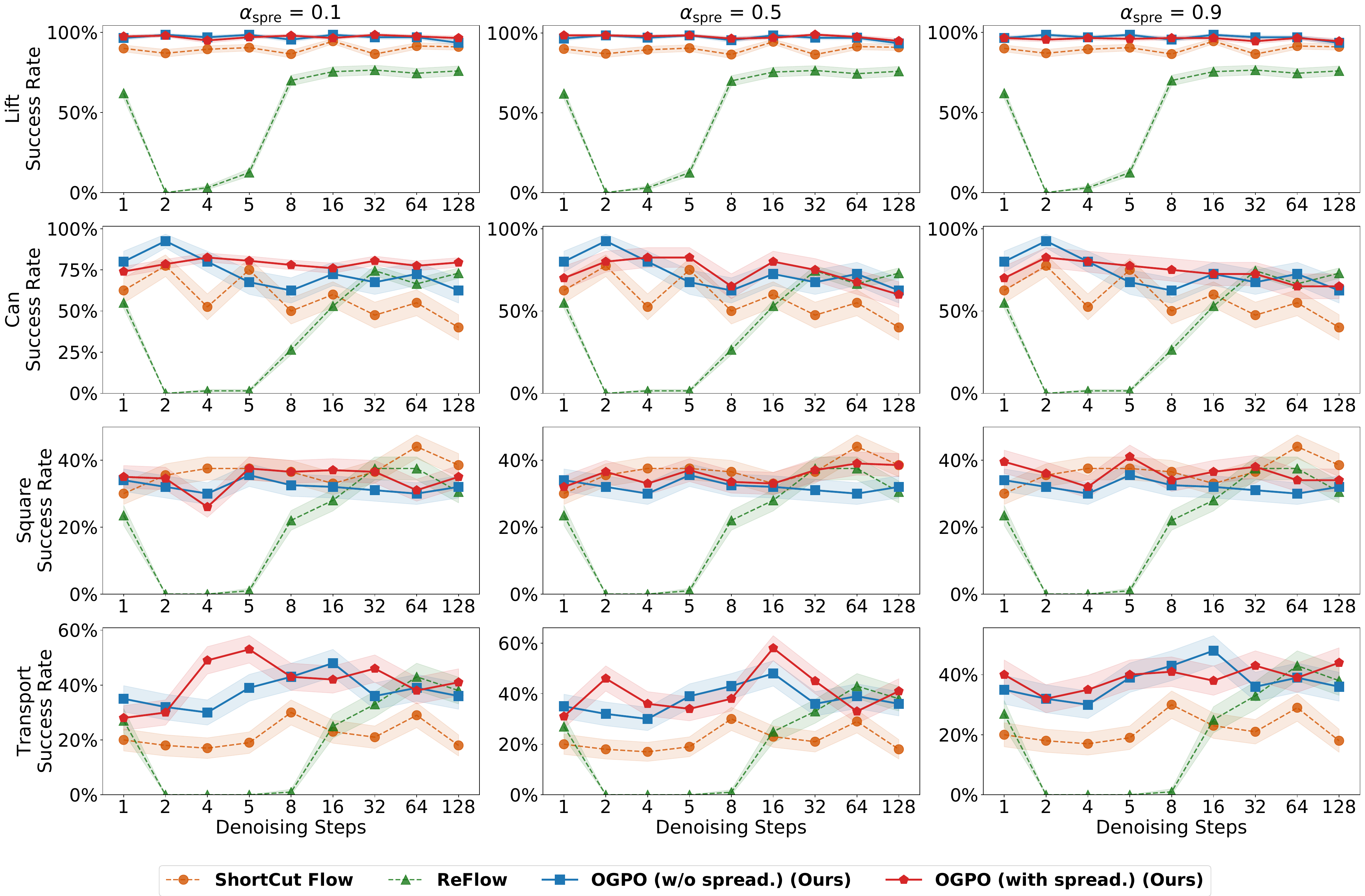}
\caption{\textbf{Stage 1 (Pre-training):} Success rate vs.\ denoising steps on a 4$\times$3 grid. Rows correspond to Lift, Can, Square, and Transport tasks with increasing difficulty, and columns correspond to $\alpha_{\text{spre}} \in \{0.1, 0.5, 0.9\}$. CVF variants, CVF and CVF+Spread, achieve near-saturated performance at 1--5 steps, while ReFlow and ShortCut require 32--128 steps. Representation spreading reduces variance on complex tasks.}
\label{fig:q2}
\end{figure}

Figure~\ref{fig:q2} reveals the effect of representation spreading across different denoising steps. ReFlow, shown as green triangles, performs poorly at low steps: on Lift, 1-step success is only $\sim$60\%, requiring 32--64 steps to approach 95\%. This indicates that learned vector fields produce highly curved trajectories with large integration errors under few-step sampling. ShortCut, shown as orange circles and designed for few-step inference, improves over ReFlow but exhibits unstable performance with limited ceiling. CVF, shown as blue squares, achieves near-saturated performance at 1--5 steps even without representation spreading. CVF+Spread, shown as red circles, further stabilizes performance, with noticeably reduced variance as evidenced by narrower shaded regions on harder tasks such as Square and Transport.

The three columns reveal the effect of $\alpha_{\text{spre}}$: representation spreading consistently improves task success rate across all tested values (0.1, 0.5, 0.9), demonstrating robustness rather than hyperparameter sensitivity. The improvement is more pronounced on complex tasks, confirming that representation spreading \textbf{prevents representation quality degradation}, which is essential for stable one-step generation and the subsequent RL fine-tuning in Stage~2.

We observe a strong linear relationship between success rate improvement and task complexity (Pearson $r = 0.924$), suggesting a simple guideline: increase $\alpha_{\text{spre}}$ for more challenging tasks.

On the additional Gym locomotion and Kitchen pre-training tasks, OGPO retained strong low-step performance. At one denoising step, it achieved 1364.1 reward on Hopper and 52.5\% completion on Kitchen-Complete, compared with 1192.6/0\% for ReFlow and 1095.3/35\% for ShortCut. OGPO also reached 3033.0 reward on Walker2d at two steps and 3235.6 reward on Humanoid at eight steps. These results extend the RoboMimic ablation to locomotion and Kitchen tasks, while the relative effect of representation spreading varies across tasks and sampling steps. Together, the ablation shows that CVF enables viable one-step policies and representation spreading improves robustness on harder tasks. The effect is most visible under the one-step budget used in the subsequent experiments.

\subsection{Stage2: Fine-Tuning Results of OGPO (RQ2)}

To evaluate our core contribution of RL fine-tuning for one-step generative policies (RQ2), we conduct experiments on both RoboMimic manipulation tasks and OpenAI Gym locomotion benchmarks.

\subsubsection{RoboMimic Fine-tuning Results}

\begin{figure}[!htbp]
\centering
\includegraphics[width=0.95\textwidth]{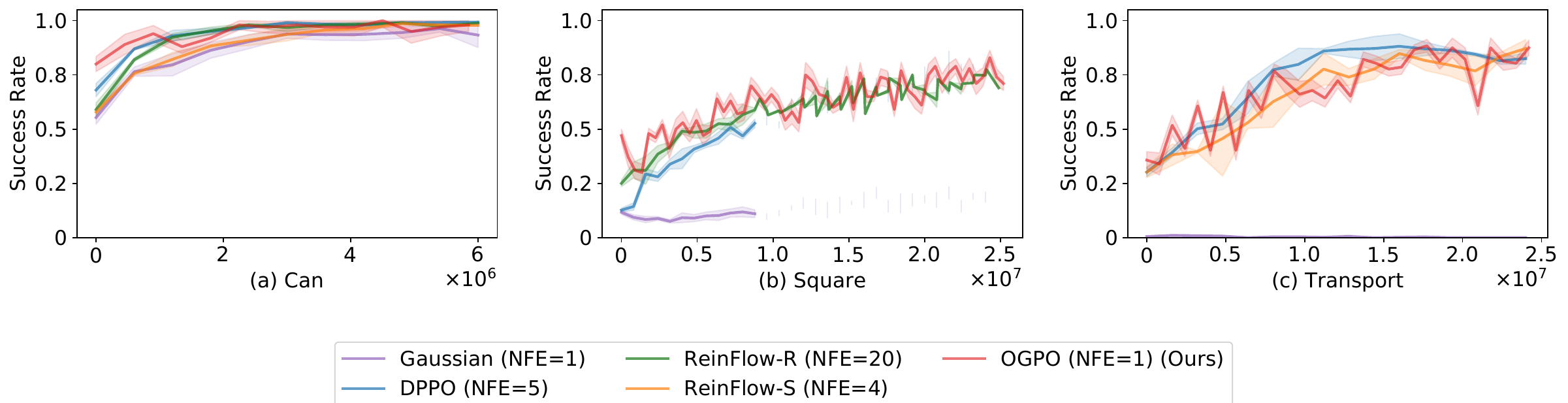}
\caption{\textbf{Stage 2 (PPO Fine-tuning):} Performance on RoboMimic manipulation tasks. CVF in blue is compared with DPPO in yellow, Gaussian in green, ReinFlow-S in red, and ReinFlow-R in purple. Lift is omitted as CVF achieves 100\% success during Stage 1 pretraining.}
\label{fig:robomimic_finetune}
\end{figure}

Figure~\ref{fig:robomimic_finetune} presents the PPO fine-tuning results on three RoboMimic manipulation tasks: Can, Square, and Transport. We omit Lift as CVF already achieves near-100\% success rate during Stage 1 pretraining. OGPO achieves competitive or superior performance compared to all baselines while requiring only 1 denoising step. On Can, OGPO achieves 100\% success rate faster than all baselines. On Square, OGPO reaches 83\% peak success rate. On Transport, OGPO achieves 88\% success rate, matching ReinFlow-S while significantly outperforming the Gaussian baseline. Across all three tasks, the RoboMimic results show that the one-step constraint does not prevent PPO from improving the pretrained policy. OGPO retains its one-step inference cost throughout fine-tuning, so these gains do not rely on restoring iterative denoising. The horizontal axis reports training environment steps, and the vertical axis reports task success rate. The curves therefore show both the rate at which PPO improves each policy and the final success level reached under the NFE specified in the legend.

\subsubsection{OpenAI Gym and Kitchen Locomotion Results}

\begin{figure}[!htbp]
\centering
\includegraphics[width=0.95\textwidth]{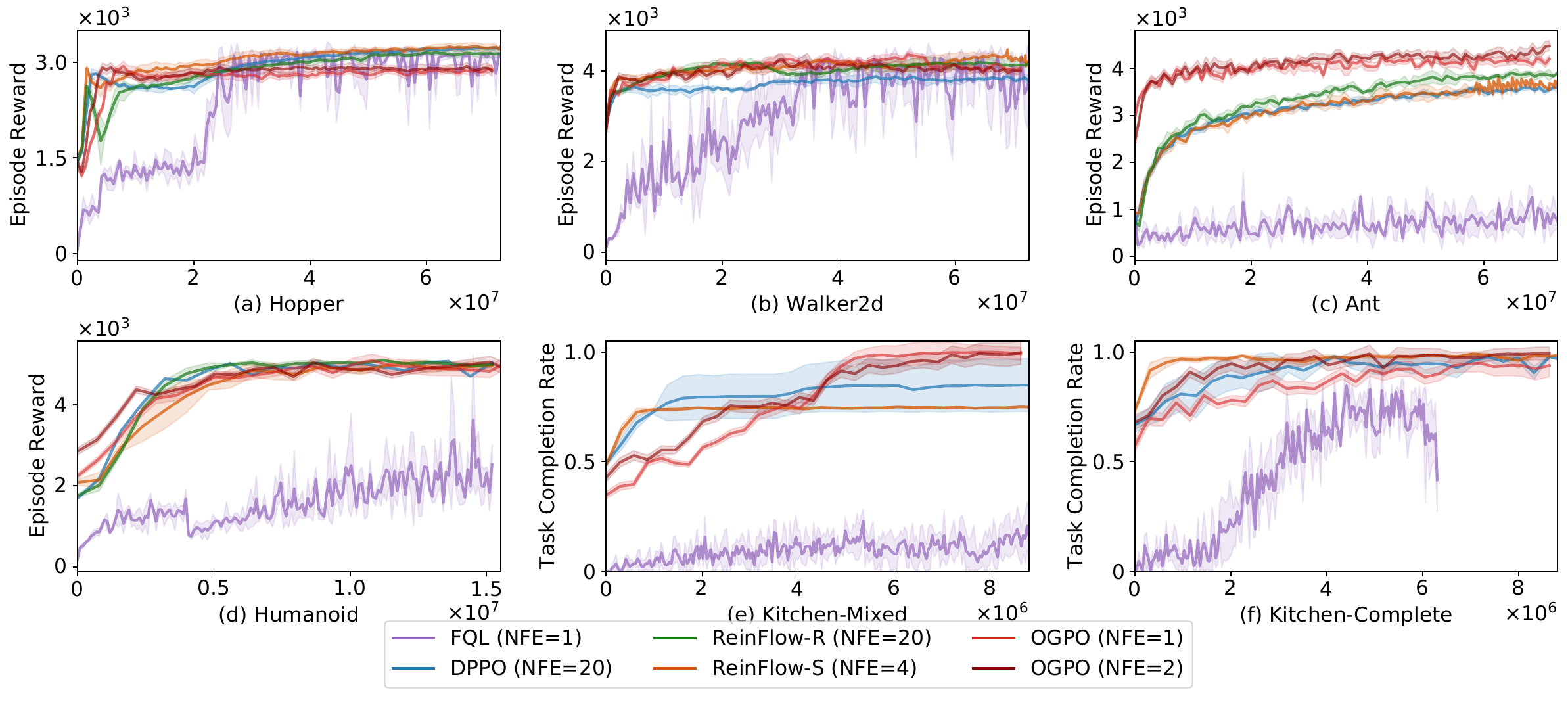}
\caption{\textbf{Stage 2 (PPO Fine-tuning):} Performance on OpenAI Gym locomotion tasks in panels a--d and Kitchen manipulation tasks in panels e--g. CVF in blue is compared with DPPO in yellow, ReinFlow-S in red, ReinFlow-R in purple, and FQL in light blue.}
\label{fig:gym_finetune}
\end{figure}

We evaluate PPO fine-tuning on four OpenAI Gym locomotion tasks, namely Hopper, Walker2d, Ant, and Humanoid, as well as two Kitchen tasks, Complete and Partial, from D4RL~\cite{d4rl2020}. As shown in Figure~\ref{fig:gym_finetune}, OGPO with 1 denoising step achieves comparable performance to multi-step baselines on Hopper, Walker2d, and Humanoid, while significantly outperforming all baselines on Ant. On Kitchen tasks, OGPO quickly reaches 100\% completion on Kitchen-Complete and maintains strong, stable performance on Kitchen-Mixed and Kitchen-Partial.

Overall, OGPO with one denoising step matches or outperforms DPPO (20 steps) and ReinFlow (4 steps), confirming that one-step generation preserves fine-tuning effectiveness. BC regularization ($\lambda_{\text{BC}} = 0.05$--$0.2$) prevents catastrophic forgetting across all tasks.

\subsection{Comparison with Existing Methods (RQ3)}

Table~\ref{tab:dataset_comparison} compares OGPO with existing one-step and multi-step methods on RoboMimic. OGPO achieves 100\% on Lift and Can, 83\% on Square, and 88\% on Transport with only 1 denoising step. OneDP achieves 78\% and 72\% on Square and Transport respectively. MP1 reaches only 35\% and 38\%. OGPO outperforms all one-step baselines by a large margin. Notably, OGPO uses only 100 trajectories yet exceeds methods trained on 300 trajectories. The Avg. column is reported only when a method has scores for all four tasks; rows with missing task results are not averaged. The simplified 100-trajectory rows provide the direct data-matched comparison between OGPO and the RL fine-tuning baselines.

\begin{table}[!htbp]
\centering
\caption{\textbf{Stage 2 (RL Fine-tuning):} Performance comparison on RoboMimic (Multi-Human data). Full dataset: 300 trajectories. Simplified: 100 trajectories. OGPO results are after Stage 2 PPO fine-tuning.}
\label{tab:dataset_comparison}
{\fontsize{9pt}{11pt}\selectfont
\begin{tabular*}{\textwidth}{@{\extracolsep{\fill}}cclrcccccc}
\toprule
\textbf{Data} & \textbf{Method} & \textbf{Venue} & \textbf{NFE} & \textbf{Dist.} & \textbf{Lift} & \textbf{Can} & \textbf{Square} & \textbf{Transport} & \textbf{Avg.} \\
\midrule
\multicolumn{10}{l}{\textit{Full Dataset (300 trajectories)}} \\
Full & LSTM-GMM & CoRL'21 \cite{robomimic2021} & - & \ding{55} & 0.93 & 0.81 & 0.59 & 0.20 & 0.63 \\
Full & IBC & CoRL'21 \cite{ibc2021} & - & \ding{55} & 0.02 & 0.01 & 0.00 & 0.00 & 0.01 \\
Full & BET & NeurIPS'22 \cite{bet2022} & - & \ding{55} & 0.99 & 0.90 & 0.43 & 0.06 & 0.60 \\
Full & DP-C & RSS'23 \cite{diffusionpolicy2023} & 100 & \ding{55} & 0.97 & 0.96 & 0.82 & 0.46 & 0.80 \\
Full & DP-T & RSS'23 \cite{diffusionpolicy2023} & 100 & \ding{55} & 1.00 & 0.94 & 0.81 & 0.35 & 0.78 \\
Full & CP & RSS'24 \cite{consistencypolicy2024} & 1 & \ding{51} & - & - & 0.65 & 0.38 & - \\
Full & OneDP-D & ICML'25 \cite{onedp2025} & 1 & \ding{51} & - & - & 0.78 & 0.71 & - \\
Full & OneDP-S & ICML'25 \cite{onedp2025} & 1 & \ding{51} & - & - & 0.77 & \textbf{0.72} & - \\
\midrule
\multicolumn{10}{l}{\textit{Simplified Dataset (100 trajectories)}} \\
Simplified & DPPO & ICLR'25 \cite{dppo2024} & 20 & \ding{55} & 1.00 & 1.00 & 0.78 & 0.80 & 0.90 \\
Simplified & ReFlow-R & NeurIPS'25 \cite{reinflow2024} & 20 & \ding{55} & 1.00 & 0.99 & 0.80 & - & - \\
Simplified & ReFlow-S & NeurIPS'25 \cite{reinflow2024} & 4 & \ding{55} & 1.00 & 0.99 & - & 0.85 & - \\
Simplified & MP1$^\dagger$ & AAAI'26 \cite{mp1_meanflow} & 1 & \ding{55} & 0.95 & 0.80 & 0.35 & 0.38 & 0.62 \\
Simplified & \textbf{OGPO (Ours)} & - & \textbf{1} & \ding{55} & \textbf{1.00} & \textbf{1.00} & \textbf{0.83} & \textbf{0.88} & \textbf{0.93} \\
\bottomrule
\end{tabular*}
}
\vspace{1mm}
\parbox{\textwidth}{\footnotesize $^\dagger$The original MP1 uses 3D point cloud input with a 255M-parameter UNet backbone, making direct comparison infeasible. We implement the core algorithm of MP1 (MeanFlow) with our 2D image input and lightweight ViT+MLP architecture for fair comparison. MP1 is better suited for comparison with 3D methods such as DP3. Results for DP-C, DP-T, CP, and OneDP are from \cite{onedp2025}. NFE = number of function evaluations. Dist.\ = knowledge distillation. Avg.\ = mean success rate over all four tasks (omitted when any task is missing).}
\end{table}

\begin{table}[!htbp]
Table~\ref{tab:model_comparison} compares model architectures and inference speed. The OGPO 1.78M-parameter ViT+MLP achieves 0.6ms inference at 1770Hz on RTX 4090. The speedup comes from two sources. First, the lightweight architecture provides 9$\times$ speedup over UNet-based CP at 1 step. Second, one-step generation provides 10$\times$ speedup over multi-step DP-DDIM. Combined, OGPO achieves 694$\times$ speedup over Diffusion Policy. OGPO's speed advantage is preserved across the two GPUs, although absolute inference latency changes with the accelerator.
\par\medskip

\centering
\caption{\textbf{Stage 2 (RL Fine-tuning): Model Architecture and Efficiency Comparison.} Speedup is relative to DP-DDPM (100 steps). All measurements use batch size 1.}
\label{tab:model_comparison}
\resizebox{\textwidth}{!}{
\begin{tabular}{cccccccccccc}
\toprule
& & & & & & \multicolumn{3}{c}{\textbf{NVIDIA RTX 4090}} & \multicolumn{3}{c}{\textbf{NVIDIA RTX 2080}} \\
\cmidrule(lr){7-9} \cmidrule(lr){10-12}
\textbf{Model} & \textbf{Vision} & \textbf{Action} & \textbf{Params} & \textbf{Size} & \textbf{Steps} & \textbf{Time} & \textbf{Freq} & \textbf{Speedup} & \textbf{Time} & \textbf{Freq} & \textbf{Speedup} \\
\midrule
\multirow{3}{*}{DP}
& \multirow{3}{*}{ResNet-18$\times$2} & \multirow{3}{*}{UNet} & \multirow{3}{*}{281.19M} & \multirow{3}{*}{$\sim$4.4GB}
& 100 (DDPM) & 391.1ms & 2.6Hz & 1$\times$ & 2007.5ms & 0.5Hz & 1$\times$ \\
& & & & & 16 (DDIM) & 63.7ms & 15.7Hz & 6$\times$ & 385.3ms & 2.6Hz & 5$\times$ \\
& & & & & 10 (DDIM) & 40.3ms & 24.8Hz & 10$\times$ & 220.4ms & 4.5Hz & 9$\times$ \\
\midrule
CP & ResNet-18$\times$2 & UNet & 284.86M & $\sim$4.4GB & 1 & 5.4ms & 187Hz & 73$\times$ & 35.5ms & 28Hz & 56$\times$ \\
\midrule
ReFlow & light ViT & MLP & 1.78M & $\sim$28MB & 20 & 8.4ms & 119Hz & 46$\times$ & 47.7ms & 21Hz & 42$\times$ \\
\midrule
ShortCut & light ViT & MLP & 1.78M & $\sim$28MB & 5 & 2.5ms & 396Hz & 155$\times$ & 12.0ms & 83Hz & 167$\times$ \\
\midrule
\multirow{3}{*}{MP1}
& \multirow{3}{*}{PointNet} & \multirow{3}{*}{UNet} & \multirow{3}{*}{255.80M} & \multirow{3}{*}{$\sim$4GB}
& 1 & 4.1ms & 244Hz & 96$\times$ & 21.4ms & 47Hz & 94$\times$ \\
& & & & & 2 & 7.7ms & 130Hz & 51$\times$ & 41.6ms & 24Hz & 48$\times$ \\
& & & & & 5 & 18.6ms & 54Hz & 21$\times$ & 102.1ms & 10Hz & 20$\times$ \\
\midrule
\multirow{3}{*}{\makecell{\textbf{OGPO}\\\textbf{(Ours)}}}
& \multirow{3}{*}{light ViT} & \multirow{3}{*}{MLP} & \multirow{3}{*}{1.78M} & \multirow{3}{*}{$\sim$28MB}
& \cellcolor{lightblue}1 & \cellcolor{lightblue}0.6ms & \cellcolor{lightblue}1770Hz & \cellcolor{lightblue}694$\times$ & \cellcolor{lightblue}2.6ms & \cellcolor{lightblue}385Hz & \cellcolor{lightblue}772$\times$ \\
& & & & & \cellcolor{lightblue}2 & \cellcolor{lightblue}1.1ms & \cellcolor{lightblue}946Hz & \cellcolor{lightblue}371$\times$ & \cellcolor{lightblue}4.9ms & \cellcolor{lightblue}204Hz & \cellcolor{lightblue}410$\times$ \\
& & & & & \cellcolor{lightblue}5 & \cellcolor{lightblue}2.6ms & \cellcolor{lightblue}386Hz & \cellcolor{lightblue}151$\times$ & \cellcolor{lightblue}12.0ms & \cellcolor{lightblue}83Hz & \cellcolor{lightblue}167$\times$ \\
\bottomrule
\end{tabular}
}
\vspace{1mm}
{\footnotesize DDPM = Denoising Diffusion Probabilistic Model. DDIM = Denoising Diffusion Implicit Model. DP = Diffusion Policy. CP = Consistency Policy.}
\end{table}

\subsection{Physical Robot Experiments}

To validate practical applicability, we deploy OGPO on a 7-Degree-of-Freedom (DOF) Franka-Emika-Panda robot equipped with an Intel RealSense D435i camera across all four RoboMimic tasks. On RTX 2080, OGPO with 1-step inference requires only 2.6ms for neural network forward pass. Including all pipeline stages, OGPO achieves 9.6ms total latency, enabling over 100Hz control frequency.

\begin{figure}[!htbp]
\centering
\includegraphics[width=0.95\textwidth]{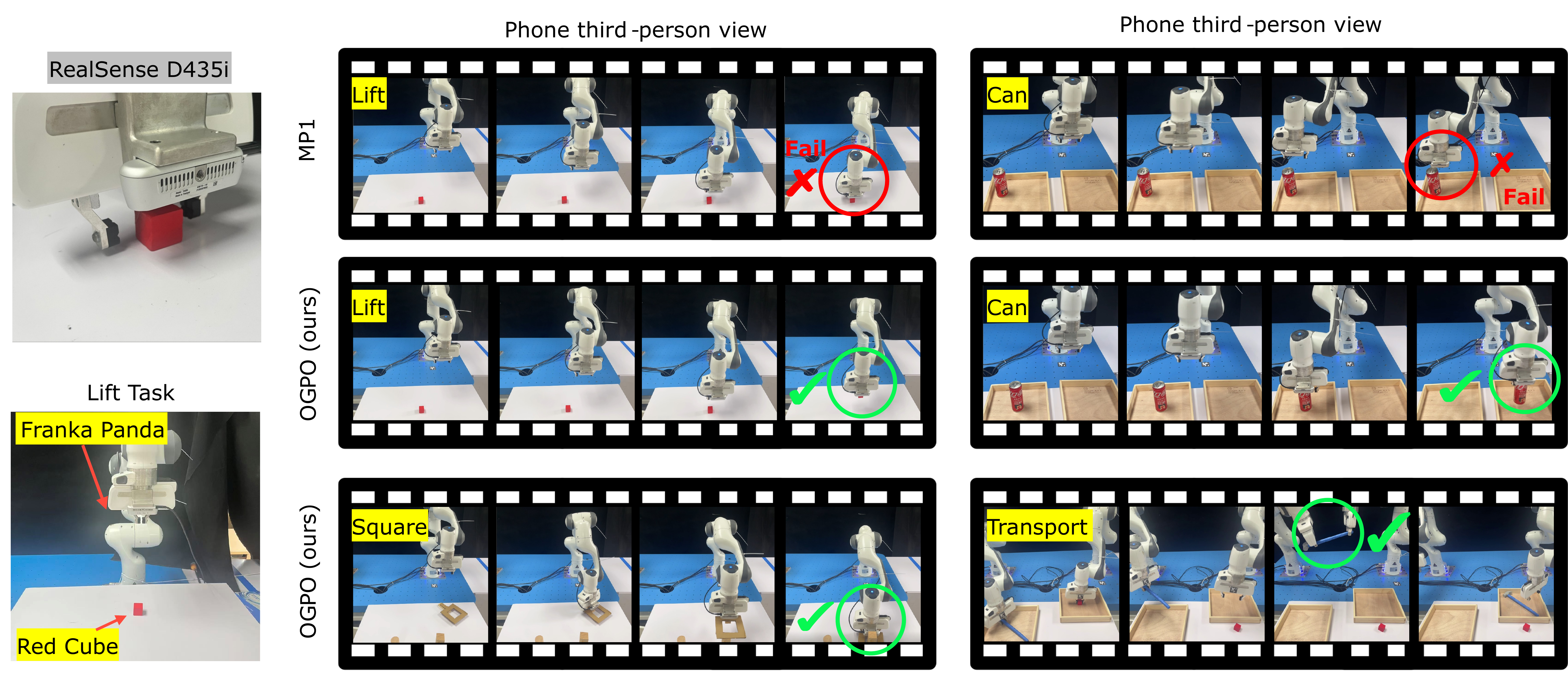}
\caption{
    \textbf{Real-world deployment of OGPO on a Franka Panda robot.} Left: Hardware setup with Intel RealSense D435i camera for visual observation. Right: Qualitative comparison between MP1 baseline (top row) and our OGPO method (rows 2-3) across four manipulation tasks. MP1 fails on Lift and Can tasks due to imprecise grasping, while OGPO successfully completes all four tasks (Lift, Can, Square, Transport), demonstrating superior sim-to-real transfer capability. Green checkmarks indicate successful task completion; red crosses indicate failure.
    }
\label{fig:real_robot}
\end{figure}

As shown in Figure~\ref{fig:real_robot}, we compare MP1 against OGPO on real hardware. We focus on MP1 as it is the only other interval-velocity-based one-step method; other baselines (DPPO, ReinFlow) are compared in simulation, and deploying multi-step methods on real robots incurs significant latency. MP1 fails on Lift and Can tasks due to imprecise grasping, as it lacks RL fine-tuning to adapt beyond the pre-trained policy. OGPO successfully completes all four tasks including Square and Transport. These results demonstrate that the OGPO two-stage design transfers to physical hardware, validating real-time control capability and sim-to-real generalization. The ablation measures how representation spreading affects one-step pre-training, while the fine-tuning results assess the policy on manipulation and locomotion tasks. The hardware experiments further show that the same one-step policy can be executed on a Franka robot without increasing the denoising budget.


\FloatBarrier
\section{Conclusion and Limitations}
\suppressfloats[t]

Real-time generative robot control faces a trilemma among inference speed, representation fidelity, and task performance. We presented OGPO, the first framework for on-policy RL fine-tuning of one-step generative robot policies, to resolve this trilemma. Stage~1 trains a Compact Velocity Field with representation spreading, producing a fast and stable one-step policy. Stage~2 performs on-policy RL fine-tuning, breaking the imitation ceiling while maintaining real-time speed. Experiments on RoboMimic, OpenAI Gym, and a physical Franka robot validate state-of-the-art performance with 5--20$\times$ inference speedup and over 120Hz control frequency.

Our contributions are threefold. (1) We identify representation quality degradation as the critical bottleneck for one-step generative policies, and show that contrastive regularization transitions from optional to structurally necessary in the one-step setting. (2) We present OGPO, the first framework to unify one-step generation with online RL fine-tuning. (3) We validate OGPO across manipulation and locomotion tasks, establishing state-of-the-art on both simulation benchmarks and physical robot deployment.

Several limitations remain. Physical deployment success rates of 60--90\% reveal a non-trivial sim-to-real gap. The current evaluation focuses on tabletop manipulation and locomotion. Performance on dexterous or contact-rich tasks remains unexplored. Scaling OGPO from the current 1.78M-parameter model to large Vision-Language-Action models remains an open direction.



\setlength{\bibsep}{2pt}
\bibliographystyle{plainnat}
\bibliography{references}
\clearpage
\includepdf[pages=-,pagecommand={\thispagestyle{empty}},addtotoc={1,section,1,Supplementary Material,ogpoappendix}]{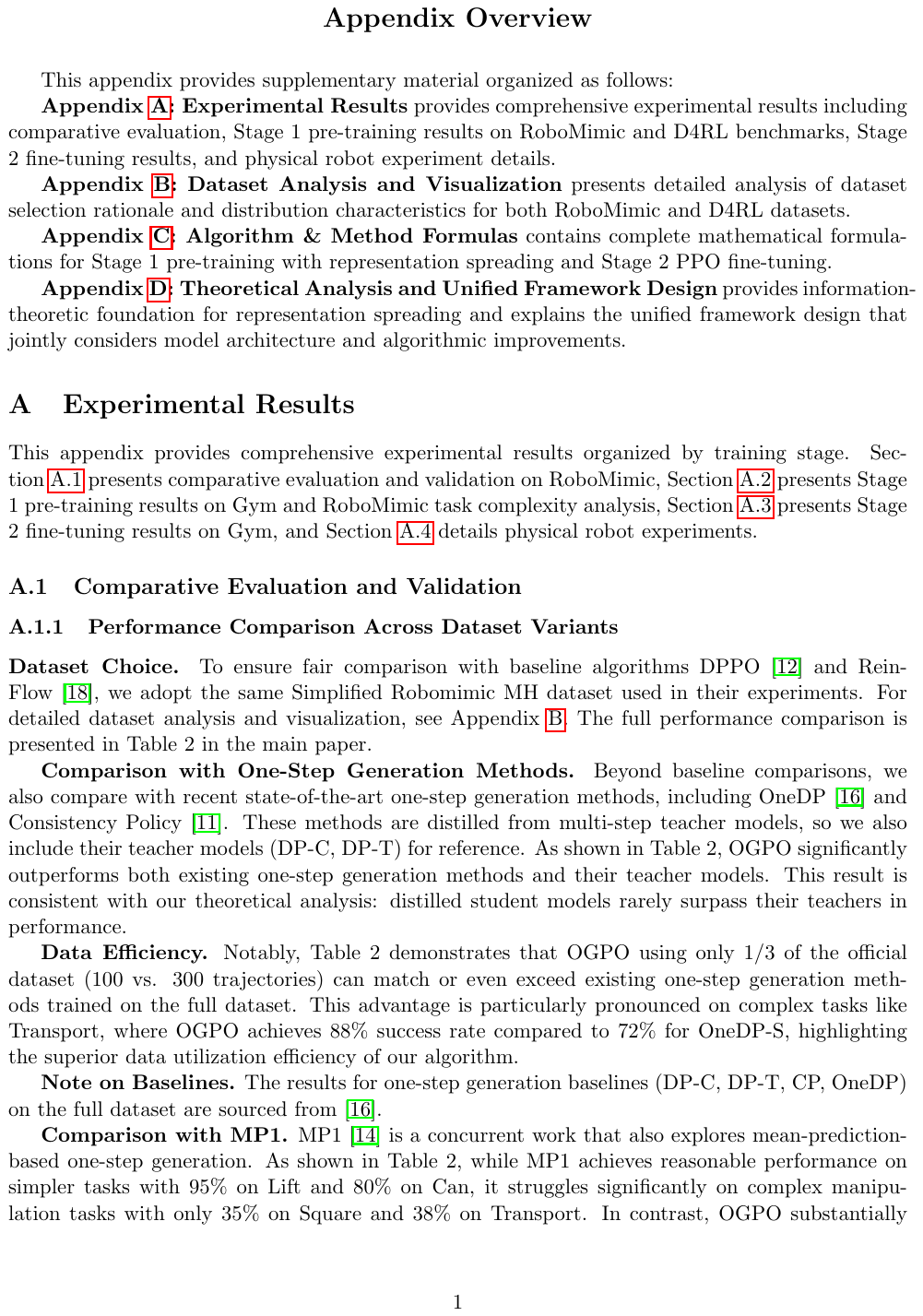}
\end{document}